\documentclass[11pt]{article}

\usepackage[utf8]{inputenc}
\usepackage[T1]{fontenc}
\usepackage{lmodern}
\usepackage[margin=1in]{geometry}
\usepackage{graphicx}
\usepackage{booktabs}
\usepackage{amsmath}
\usepackage{microtype}
\usepackage[font=small,labelfont=bf]{caption}
\usepackage[hidelinks]{hyperref}

\graphicspath{{figures/}}
\newcommand{\x}{\ensuremath{\times}}

\title{\textbf{Cortex: Compact Behavior Cloning for Quake\\with Frozen Visual Features}}
\author{Dzmitry Malyshau\\[-1pt]
\small \href{https://orcid.org/0009-0005-6410-4276}{ORCID 0009-0005-6410-4276}}
\date{%
August 2026\\[4pt]
\small Code: \href{https://github.com/kvark/cortex-actor/tree/b4de4f66420df2c408ec42b5c01c91a088d8b63d}{\texttt{cortex-actor@b4de4f6}} \;\textperiodcentered\;
Weights: \href{https://huggingface.co/mad-bot/cortex}{\texttt{mad-bot/cortex}} \;\textperiodcentered\;
Video: \href{https://youtu.be/Ou9NAmFoCOM}{\texttt{youtu.be/Ou9NAmFoCOM}}}

\begin{document}
\maketitle

\begin{abstract}
We study how far a deliberately simple behavioral-cloning policy can progress in a visually rich first-person game before adding reinforcement learning or explicit memory. \textbf{Cortex} has 10.98M trainable parameters in a six-layer transformer over a frozen 28.7M-parameter DINOv3 ViT-S+/16 encoder. It is trained on the Quake subset of the public Pixels2Play corpus: 6{,}849 recordings (${\sim}474.7$ hours), represented as 17.09M cached 10\,Hz decision frames with ground-truth keyboard and mouse actions. The complete training run is one sampled epoch---517{,}048 four-frame windows, or 3.27\% of all valid windows in the training split---requiring 32{,}320 optimizer updates and 3.3 minutes of policy-head optimization on one RTX 5080; evaluation uses its 30{,}000-update checkpoint. This timing excludes one-time feature extraction. In two independent batches of 20 stochastic, 120-second E1M1 episodes, Cortex completes the level 0/20 times in each batch, reaches pose-based proxies for the opening door, button room, and gate descent in 20/20, has route-waypoint medians 5 and 6 (maximum 9), and records at least one kill in 19/20 episodes in each batch. Under the same time-controlled environment, released P2P-150M and NitroGen checkpoints each complete 0/5 matched-duration episodes and have route median 1. These reference comparisons are limited by small samples, different native action and observation interfaces, and a custom gamepad-to-keyboard adapter for NitroGen; they measure the evaluated releases under this harness, not generalist ability in the aggregate. Additional-map and mid-map-start experiments are mixed. Controlled ablations show that denser visual tokens improve combat and survival but not route reliability, while longer optimization and naive action history improve offline metrics without consistently improving play. The remaining failures are consistent with covariate shift and motivate targeted corrective data. We release the compact policy implementation, checkpoint, and a representative rollout.
\end{abstract}

\section{Introduction}\label{sec:intro}

Large gaming policies demonstrate impressive breadth. NitroGen~\cite{nitrogen} trains a roughly 500M-parameter gamepad policy on 40{,}000 hours across more than 1{,}000 games; Pixels2Play (P2P)~\cite{p2p} releases keyboard-and-mouse policies trained on 8{,}300+ hours. Breadth and depth are different questions, however. Before adding reinforcement learning, navigation memory, or goal conditioning, we ask what a compact specialist can learn from ordinary supervised imitation on one target game.

Quake E1M1 is a useful stress test. Its first minute combines locomotion, door interaction, a wall button, combat, vertical transitions, and recovery from collisions. Completion is a long-horizon binary outcome, but intermediate progress can be measured from an engine pose stream. The level therefore exposes both useful imitation and the familiar behavioral-cloning failure mode: small errors lead to observations absent or rare in the demonstration distribution.

We factor generality into a frozen visual representation and a small learned controller. Cortex uses DINOv3 features, but the policy itself receives no privileged state, map, pose, goal, or Quake-specific behavior rule. We train it on the Quake recordings in P2P's public corpus. Sharing data provenance makes P2P an informative reference, but it does \emph{not} isolate architecture: P2P is trained with a different multi-game distribution, objective, visual encoder, action decoder, temporal context, and correction data. NitroGen is a still looser reference because its native action space is a gamepad and our Quake mapping is an adapter choice.

Our contributions are:
\begin{itemize}
\item a compact baseline with 10.98M trainable policy parameters over a frozen visual encoder, direct held-state and mouse prediction, and no learned action history;
\item an exact account of source frames, cached decision frames, sample coverage, optimization time, and parameter accounting;
\item an engine-observed evaluation with authoritative completion and death signals, pose-derived intermediate progress, two $N{=}20$ Cortex batches, and matched-duration reference batches for two released generalists;
\item controlled, retained ablations of spatial density, temporal sampling, optimization exposure, action history, and recovery-weighted data selection; and
\item public policy code, the exact evaluated weights, and a representative gameplay video.
\end{itemize}

\section{Related work}\label{sec:related}

\paragraph{Generalist gaming policies.} NitroGen~\cite{nitrogen} couples a SigLIP2-L~\cite{siglip2} vision tower with a flow-matching action decoder. Its 40{,}000-hour dataset is built from internet videos that display controller overlays: the overlay is localized and parsed with a trained segmentation/classification model, followed by action-density filtering. It is therefore inaccurate to describe NitroGen's labels as outputs of a gameplay inverse-dynamics model. The paper trains with one context frame and 16-action chunks; the released checkpoint evaluated here serializes an 18-action horizon. P2P~\cite{p2p} is a decoder-only transformer over an EfficientNet-B0~\cite{efficientnet} visual token with a 200-frame history and an autoregressive keyboard/mouse decoder. Its public corpus contains 8{,}300+ hours of recorded human inputs. P2P explicitly conditions its backbone on past ground-truth action tokens during training and mixes in human correction trajectories (less than 1\% of annotated data), two details directly relevant to recovery. SIMA~2~\cite{sima2} combines Gemini with a task-conditioned embodied agent that follows language and image goals, transfers to held-out 3D worlds, and improves from self-generated experience. It is relevant to the long-term goal of broadly capable game agents, but differs from our unconditional low-level imitation setting and was announced as a \href{https://deepmind.google/blog/sima-2-an-agent-that-plays-reasons-and-learns-with-you-in-virtual-3d-worlds/}{limited research preview}; no compatible released checkpoint was available for this harness.

\paragraph{FPS agents.} ViZDoom~\cite{vizdoom} established Doom as a platform for visual RL; population-based reinforcement learning reached human-level capture-the-flag play in Quake III Arena~\cite{ftw}; and large-scale behavioral cloning produced competent Counter-Strike deathmatch behavior~\cite{csgo}. Relative to the short, combat-centric ViZDoom scenarios commonly used as benchmarks, E1M1 requires a convoluted multi-level route, doors and a wall button, lifts and other vertical transitions, vertical aiming, and recovery amid low-fidelity, visually repetitive textures. This characterizes the task used here, not Doom as a whole, which can also support navigation-rich scenarios.

\paragraph{Frozen-feature control.} Frozen pretrained vision models can be competitive representations for control~\cite{pvr}, and R3M~\cite{r3m} and VC-1~\cite{vc1} study this pattern in robot learning. The DINO family~\cite{dino,dinov2,dinov3} provides dense self-supervised features. Cortex trains on cached and deploys on streamed DINOv3 activations; the complete deployed system still includes the frozen encoder.

\paragraph{Imitation learning.} ALVINN~\cite{alvinn} is an early neural behavioral-cloning system; DAgger~\cite{dagger} formalizes compounding error and introduces interactive data aggregation. P2P's correction trajectories motivate a related corrective-data direction at small scale. Robomimic reports that validation loss can be a poor predictor of closed-loop manipulation success~\cite{robomimic}. Action-sequence prediction in ACT~\cite{act} and Diffusion Policy~\cite{diffusionpolicy} motivates coherent multi-step decoders, but the baseline studied here deliberately retains a simpler per-decision head.

\section{Architecture}\label{sec:arch}

Cortex is deliberately minimal. Per 100\,ms policy decision:
\begin{itemize}
\item \textbf{Vision.} A frozen DINOv3 ViT-S+/16~\cite{dinov3} encodes a 640\x400 frame into a CLS token and a 25\x40 patch grid. The policy consumes CLS and an ordered 5\x8 uniform sample of the patch grid: 41 tokens of dimension 384 per frame.
\item \textbf{Trunk.} Four frames spanning 300\,ms yield 164 tokens with learned spatial and temporal embeddings. A six-layer, 384-dimensional, six-head bidirectional transformer (feed-forward dimension 1536) produces a summary at the last frame's CLS position.
\item \textbf{Heads.} One linear head predicts 36 independent held-state logits (33 keys and three mouse buttons) with binary cross-entropy. Two linear heads predict tanh-squashed relative mouse dx/dy with smooth-L1 loss and deploy scales of $(\pm500,\pm250)$ counts.
\item \textbf{Execution.} Held states are sampled independently at temperature 1, masked by the game adapter's legal action schema, and differenced against the previously executed state to generate press/release events. The policy receives no previous action, pose, map, task text, or privileged observer state.
\end{itemize}

\begin{figure}[ht]
\centering
\includegraphics[width=\linewidth]{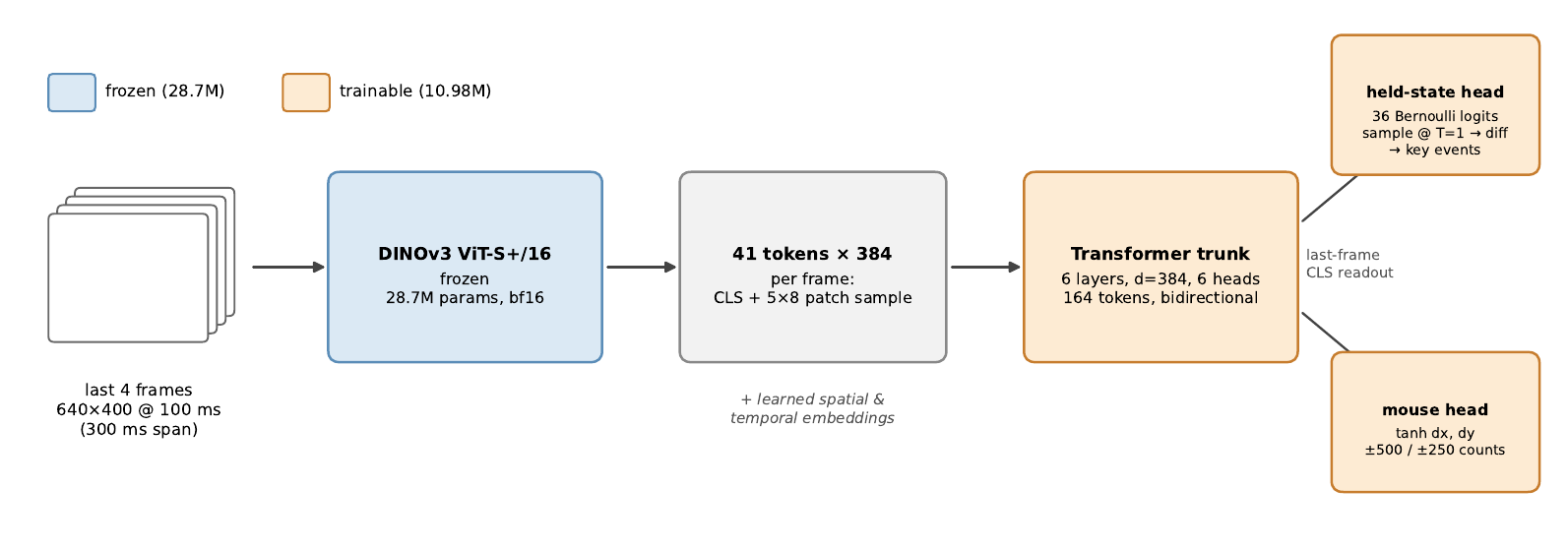}
\caption{Cortex architecture. The DINOv3 encoder is frozen; the transformer and action heads contain 10.98M trainable parameters.}
\label{fig:arch}
\end{figure}

\paragraph{Parameter accounting.} The released state dictionary contains exactly 10{,}975{,}142 policy parameters. DINOv3 ViT-S+/16 adds approximately 28.7M frozen parameters, so the pixel-to-action system executes about 39.7M parameters per decision. P2P-150M and NitroGen's roughly 493M parameters include their visual stacks; these size labels are not matched compute measurements. P2P initializes an EfficientNet encoder from ImageNet and trains it jointly, while NitroGen fine-tunes its SigLIP2 tower; Cortex does not update DINOv3. We report both 10.98M trainable and 39.7M total, and make no claim that parameter count alone causes the measured behavior.

\begin{table}[ht]
\centering
\small
\begin{tabular}{ll}
\toprule
Component & Value \\
\midrule
Vision encoder & DINOv3 ViT-S+/16, frozen, ${\sim}28.7$M, 640\x400 \\
Tokens per frame & 1 CLS + 40 sampled patches, 384-d \\
Context & 4 frames / 300\,ms \\
Policy trunk & 6 layers, $d{=}384$, 6 heads, FF 1536 \\
Action heads & 36 held-state logits + continuous mouse dx/dy \\
Decision rate & 10\,Hz simulated time \\
Trainable / total parameters & 10.98M / ${\sim}39.7$M \\
Sampled epoch & 32{,}320 updates, 517{,}048 examples, maximum batch 16 \\
Evaluated checkpoint & update 30{,}000 \\
Measured throughput & 2{,}601 examples/s; 198.8\,s on RTX 5080 \\
\bottomrule
\end{tabular}
\caption{Architecture and policy-optimization summary. Feature extraction is excluded from the 198.8-second timing.}
\label{tab:arch}
\end{table}

\section{Data and training}\label{sec:data}

We use the Quake subset of the public Pixels2Play corpus (\texttt{elefantai/p2p-full-data}), with recorded human keyboard and mouse input. The packed index contains \textbf{6{,}849 recordings} and approximately \textbf{474.7 hours}. At the source rate this corresponds to approximately 34.18M frames at 20\,fps. We retain every second source frame for the 10\,Hz policy, producing exactly \textbf{17{,}087{,}846 cached decision frames}.

Observation $i$ is paired with the action interval beginning at $i+1$. The source rate, stride 2, and offset 1 are recorded in the packed index and checkpoint contract. The training split contains 129{,}262 chunks and 15{,}828{,}466 valid four-frame windows. We select four deterministic, distinct windows per chunk, or 517{,}048 examples (3.27\% of the valid-window universe). Thus ``one epoch'' means one pass over this selected sample, not one pass over every possible window.

We hold out 342 whole recordings, yielding 844{,}831 valid validation windows before evaluation subsampling. Frames from one recording never cross splits. The production sample uses no augmentation, class reweighting, route labels, door oversampling, or privileged game state. At 2{,}601 examples/s, the complete sampled epoch takes 198.8 seconds; the evaluated snapshot is saved at update 30{,}000. This excludes DINOv3 extraction, packing, validation, and rollouts. At a measured selective-encoder throughput of approximately 207 frames/s, encoding all 17.09M retained decision frames is projected to take approximately 23 GPU-hours; this is not an end-to-end timing of a complete extraction run.

\section{Evaluation protocol}\label{sec:protocol}

All systems play the same vkQuake build through the same capture path and virtual input device. A read-only engine channel exports player pose, health, kills, and client intermission state without changing game rules or physics. The environment advances virtual time in 1/60-second substeps and uses a 61.2\,Hz render ceiling for scheduling headroom. This describes action/observation granularity, not every internal engine subsystem. Cortex holds one decision for six substeps (100\,ms), P2P for three (50\,ms), and NitroGen's queued gamepad actions are applied one per substep.

\paragraph{Time control.} The game waits while a policy computes, so cadence is simulated rather than wall-clock time. This removes stale-frame and missed-decision effects but does not measure real-time deployment throughput. P2P targets real-time 20\,Hz inference and reports an RTX 5090 for policy inference plus an RTX 5080 for rendering. NitroGen's own simulator also freezes game time during inference~\cite{nitrogen}; our timing model therefore aligns with NitroGen rather than departing from it.

\paragraph{Reference implementations.} Both references use released checkpoints and upstream model-side preprocessing/decoding, wrapped by compatibility glue and our game adapter.
\begin{itemize}
\item \textbf{P2P-150M}: released step-500k checkpoint, upstream KV-cache state, 192\x192 Hamming preprocessing, autoregressive action decoding at temperature 1.0, truncated-normal mouse-bin dequantization, 20\,Hz, and Quake sensitivity 3.5 from P2P's documented setup. Its rolling context is reset at episode boundaries but not periodically within an episode.
\item \textbf{NitroGen}: released checkpoint and processor, 256\x256 input, bf16, and the checkpoint's serialized 18-action horizon, applied at 60\,Hz. NitroGen emits gamepad controls, so we use a custom mapping: left stick to thresholded WASD, right stick to mouse counts, and SOUTH to fire (Appendix~\ref{app:mapping}). No official NitroGen-to-Quake binding was released, and we bind no jump button. Results are adapter-dependent.
\end{itemize}
Cortex and NitroGen use Quake sensitivity 6.0; P2P retains its documented 3.5. The systems therefore share the environment, observer, capture, and input-injection mechanism, but retain model-native preprocessing, cadence, action representation, and sensitivity.

\paragraph{Outcomes and progress.}
\begin{itemize}
\item \textbf{Completion} is authoritative: success requires the engine's level-to-intermission transition. Death is health $\leq0$ from the same observer.
\item \textbf{Route waypoint} is a heuristic: the largest index among 15 reference points (0--14) whose 128-unit radius is touched at any time. Earlier points are not prerequisites. ``Opening sequence'' below is shorthand for the pose reaching the door, button-room, and gate-descent regions; it is not a direct button-event log.
\item \textbf{Batches}: each Cortex batch has 20 episodes (four lanes and five policy seeds), 120 simulated seconds, stochastic decoding, and no episode selection. Each reference has five matched-duration episodes. Videos were recorded and inspected; the retention policy later removed some raw videos after preserving contact sheets, telemetry, summaries, and manifests.
\item \textbf{Statistics}: binomial rates carry Wilson 95\% intervals. We report every episode but do not run significance tests on route or combat metrics. Five-episode reference batches are descriptive and cannot establish a broad ranking.
\end{itemize}

\section{Inference performance on one RTX 5080}\label{sec:latency}

We measure the four evaluated pixel-to-action stacks on the same RTX 5080, in isolation, at batch size one. A retained E1M1 frame is converted once to each model's native input; capture, video decoding, model-external CPU image preprocessing, game execution, and input injection are outside the timed region. The input tensor is already GPU-resident. We use each evaluator's eager, model-native precision path, warm up Cortex for 20 calls, NitroGen for 10, and P2P for 210 so its 200-frame rolling KV cache is full, then record 100 calls with CUDA events. The vision interval brackets each implementation's native visual module. ``Policy/action'' is the same-call residual and includes downstream tensor operations, policy inference, and native action generation. These are measurements of the evaluated implementations, not hardware-independent architectural lower bounds.

\begin{figure}[ht]
\centering
\includegraphics[width=0.95\linewidth]{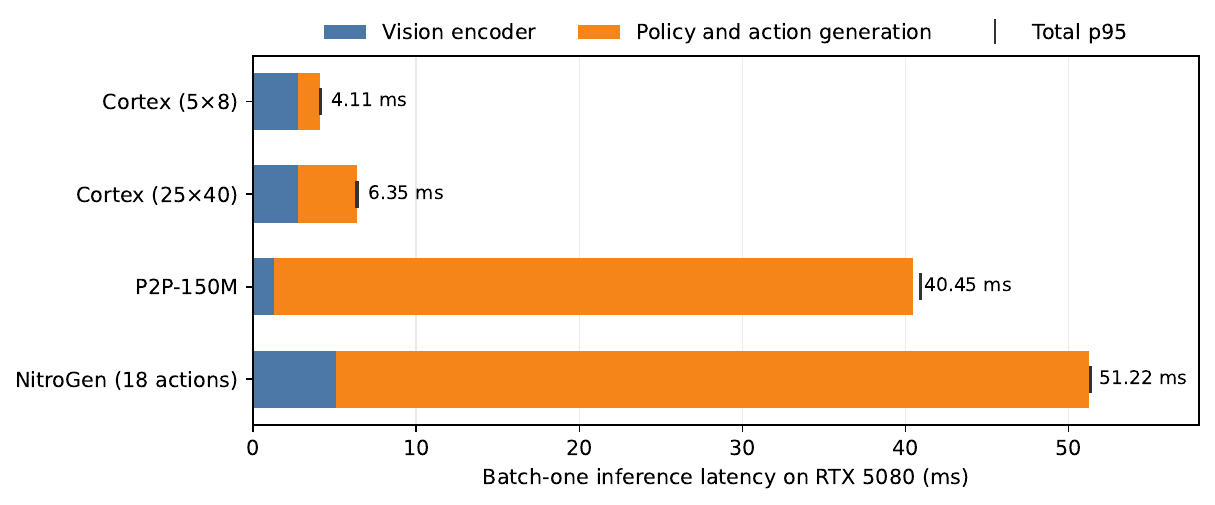}
\caption{Median batch-one inference latency on one RTX 5080. Bars use native component boundaries and native outputs; the black mark is total p95. NitroGen produces an 18-action chunk per call, whereas Cortex and P2P produce one decision.}
\label{fig:latency}
\end{figure}

\begin{table}[ht]
\centering
\small
\setlength{\tabcolsep}{4pt}
\begin{tabular}{lrrrrl}
\toprule
System & Vision p50 & Policy/action p50 & Total p50 & Total p95 & Native output \\
\midrule
Cortex 5\x8 & 2.751 & 1.363 & \textbf{4.114} & 4.141 & 1 decision \\
Cortex 25\x40 & 2.761 & 3.592 & 6.352 & 6.378 & 1 decision \\
P2P-150M & 1.258 & 39.190 & 40.448 & 40.909 & 1 decision \\
NitroGen & 5.100 & 46.124 & 51.224 & 51.335 & 18 actions \\
\bottomrule
\end{tabular}
\caption{RTX 5080 inference latency in milliseconds over 100 measured calls per system. Component medians are computed from paired per-call intervals; rounding accounts for any difference from the total.}
\label{tab:latency}
\end{table}

Compact Cortex takes 4.11\,ms at p50. Consuming all 25\x40 DINO patches raises its policy segment by only 2.23\,ms and total latency to 6.35\,ms. Thus the full-grid experiment's approximately 30-fold training-throughput penalty is primarily a training and packed-data cost, not a comparable deployment penalty at batch one. P2P takes 40.45\,ms per 20\,Hz decision in this eager steady-state implementation. NitroGen takes 51.22\,ms per 18-action chunk, or 2.85\,ms per queued 60\,Hz action when amortized, although chunked and single-decision policies are not equivalent control interfaces. All p95 values fit their native simulated-time coverage (100\,ms Cortex, 50\,ms P2P, 300\,ms NitroGen), although time control imposes no wall-clock deadline.

\section{Results}\label{sec:results}

\subsection{E1M1 from a fresh spawn}\label{sec:main}

\begin{figure}[ht]
\centering
\includegraphics[width=0.9\linewidth]{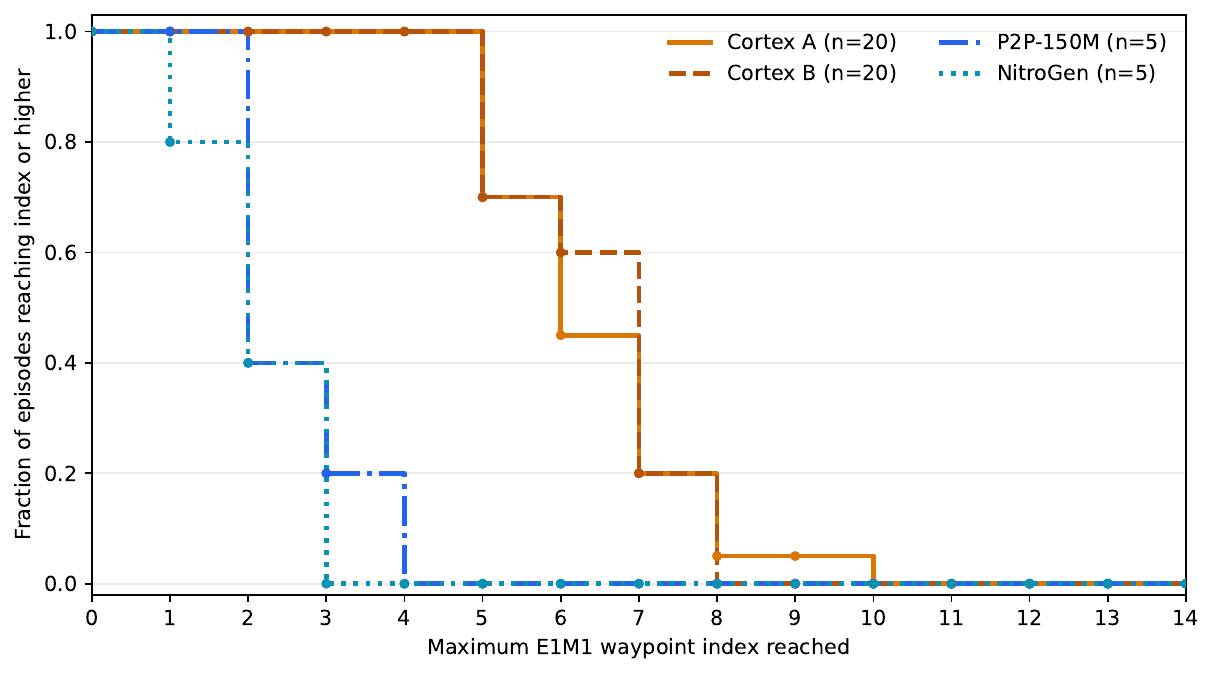}
\caption{Fraction of episodes whose pose reaches waypoint $k$ or any higher-index waypoint on E1M1. Both Cortex batches have $n{=}20$; each reference has $n{=}5$. The route score is heuristic and order-independent.}
\label{fig:survival}
\end{figure}

\begin{table}[ht]
\centering
\scriptsize
\setlength{\tabcolsep}{3pt}
\begin{tabular}{lcccc}
\toprule
 & Cortex ($N{=}20$) & Replication ($N{=}20$) & P2P-150M ($N{=}5$) & NitroGen ($N{=}5$) \\
\midrule
Trainable parameters & 10.98M (39.7M total) & --- & ${\sim}150$M & ${\sim}493$M \\
Training data & 474.7\,h Quake & --- & 8{,}300+\,h multi-game & 40{,}000\,h multi-game \\
Completion & 0/20 & 0/20 & 0/5 & 0/5 \\
Door/room/descent pose proxies & \textbf{20/20 each} & \textbf{20/20 each} & 2/5 / 1/5 / 0/5 & 2/5 / 0/5 / 0/5 \\
Route median (maximum) & 5 (9) & \textbf{6 (7)} & 1 (3) & 1 (2) \\
Episodes with a kill & 19/20 & 19/20 & 2/5 & 0/5 \\
Total kills & 32 & 28 & 2 & 0 \\
Deaths & 15/20 & 18/20 & 0/5 & 2/5 \\
\bottomrule
\end{tabular}
\caption{Fresh-spawn E1M1 measurements, 120 simulated seconds per episode. Different native interfaces and unequal sample sizes limit cross-system inference.}
\label{tab:main}
\end{table}

For completion, 0/20 has a Wilson 95\% interval of [0\%, 16.1\%], while 0/5 has [0\%, 43.4\%]. The opening-region rate 20/20 has [83.9\%, 100\%]. Pooling the two fresh-seed Cortex batches descriptively gives 0/40 completions; this does not convert the waypoint heuristic into success.

In the production batch, median maximum displacement is 1{,}481 units (maximum 2{,}526), median path length is 10{,}520, and 19/20 episodes record a kill. The replication again records a kill in 19/20. Reaching the gate descent strongly suggests that the opening interaction succeeded, but telemetry has no direct button-press event, so we do not report 20/20 button presses. A representative episode is available at \url{https://youtu.be/Ou9NAmFoCOM}.

The matched 120-second P2P episodes have route median 1 (maximum 3), displacement 914--1{,}064, two total kills, and stationary stretches up to 34.5 seconds. NitroGen has route median 1 (maximum 2), maximum displacement 1{,}047, zero kills, and two deaths. In these ten episodes, extending the earlier 60-second screens does not yield deeper progress. That observation rules out the shorter cutoff for these traces, not an intrinsic ceiling for either released model.

\begin{table}[ht]
\centering
\small
\begin{tabular}{lcccccc}
\toprule
System, ep. & Route & Max displ. & Path & Longest stationary & Kills & Outcome \\
\midrule
P2P 0 & 1 & 914 & 2{,}190 & 11.4\,s & 0 & time limit \\
P2P 1 & 1 & 1{,}045 & 5{,}127 & 21.9\,s & 0 & time limit \\
P2P 2 & 3 & 1{,}063 & 5{,}172 & 14.9\,s & 1 & time limit \\
P2P 3 & 1 & 919 & 4{,}673 & 6.8\,s & 0 & time limit \\
P2P 4 & 2 & 1{,}045 & 4{,}986 & 34.5\,s & 1 & time limit \\
\midrule
NitroGen 0 & 2 & 1{,}044 & 4{,}232 & 4.3\,s & 0 & died @ 59\,s \\
NitroGen 1 & 1 & 1{,}043 & 9{,}170 & 4.6\,s & 0 & time limit \\
NitroGen 2 & 2 & 1{,}047 & 10{,}983 & 5.6\,s & 0 & died @ 89\,s \\
NitroGen 3 & 0 & 299 & 4{,}216 & 19.3\,s & 0 & time limit \\
NitroGen 4 & 1 & 759 & 11{,}416 & 3.9\,s & 0 & time limit \\
\bottomrule
\end{tabular}
\caption{Per-episode measurements for the matched-duration reference batches.}
\label{tab:baseline-eps}
\end{table}

\subsection{Exploratory additional-map evaluation}\label{sec:maps}

We ran all systems on E1M2 and E1M3 for five 120-second episodes each. The policy has no map identifier or map-specific behavior rule, but five episodes cannot establish map generalization. Because the route ladder is E1M1-specific, we report displacement, combat, and survival. One P2P episode on each map and one NitroGen episode on E1M3 ended in an environment truncation; all remain in the $N{=}5$ batches. Chord summaries use every episode with a valid pose sample, while survival uses the observed duration up to death, time limit, or truncation.

\begin{table}[ht]
\centering
\small
\begin{tabular}{llcccc}
\toprule
Map & System & Chord median (best) & Kills & Died & Median survival \\
\midrule
E1M2 & Cortex & \textbf{791 (2{,}008)} & \textbf{6} & 5/5 & 25\,s \\
E1M2 & P2P-150M & 704 (791) & 4 & 3/5 & 37\,s \\
E1M2 & NitroGen & 679 (797) & 1 & 5/5 & 26\,s \\
\midrule
E1M3 & Cortex & \textbf{706 (1{,}126)} & 12 & 5/5 & 14\,s \\
E1M3 & P2P-150M & 658 (706) & \textbf{15} & 3/5 & 46\,s \\
E1M3 & NitroGen & 390 (529) & 0 & 4/5 & 56\,s \\
\bottomrule
\end{tabular}
\caption{Exploratory additional-map measurements ($N{=}5$ per system and map).}
\label{tab:maps}
\end{table}

Cortex has the largest median displacement on both maps and more kills on E1M2; P2P has more kills and longer survival on E1M3. No evaluated episode completes either map. The small batches and three environment truncations make these results especially preliminary; the mixed outcomes do not establish map generalization.

\subsection{Exploratory shared mid-map starts}\label{sec:spawns}

Two E1M1 save states, at 30 and 60 seconds into a reference trajectory, are loaded identically for every system. State B lies past the opening descent. Cortex/state A retains four episodes after a harness incident; every other cell has five. Each save already contains one kill; Table~\ref{tab:spawns} reports only kills added during the evaluated continuation.

\begin{center}
\begin{minipage}{\textwidth}
\centering
\small
\setlength{\tabcolsep}{4pt}
\begin{tabular}{llccccc}
\toprule
Start & System & Sorted route indices & Chord median (best) & New kills & Died & Median survival \\
\midrule
A & Cortex & [5,5,6,6] & \textbf{919 (950)} & 1 & 3/4 & 55\,s \\
A & P2P-150M & [0,4,5,6,7] & 484 (\textbf{1{,}375}) & 4 & 2/5 & \textbf{120\,s} \\
A & NitroGen & [5,5,5,6,6] & 661 (731) & 2 & 5/5 & 25\,s \\
\midrule
B & Cortex & [5,5,6,6,7] & 628 (716) & 2 & 5/5 & 32\,s \\
B & P2P-150M & \textbf{[6,6,7,7,7]} & 652 (966) & \textbf{5} & 3/5 & 26\,s \\
B & NitroGen & [5,5,5,5,6] & 230 (406) & 1 & 5/5 & 11\,s \\
\bottomrule
\end{tabular}
\captionof{table}{Exploratory shared-state starts. Waypoint indices are order-independent and include the starting region.}
\label{tab:spawns}
\end{minipage}
\end{center}

All systems act nontrivially from at least one mid-map state. Cortex has the largest median displacement from A; P2P is most consistent from B and most survivable from A; NitroGen generally advances least and dies fastest. These measurements are diagnostic rather than a robustness claim.

\section{Controlled ablations and failure analysis}\label{sec:negative}

Table~\ref{tab:negative} includes only comparisons with retained checkpoints, manifests, and rollout summaries. Most small screens use four episodes and should be interpreted as elimination tests, not precise effect estimates.

\begin{table}[ht]
\centering
\footnotesize
\setlength{\tabcolsep}{3.5pt}
\begin{tabular}{p{0.24\linewidth}p{0.14\linewidth}p{0.53\linewidth}}
\toprule
Change & Live sample & Result relative to compact production \\
\midrule
5\x8 $\to$ 10\x16 patches & $N{=}20$ & Offline imitation improves; route median 5$\to$4, descent proxy 20/20$\to$14/20, kill incidence 19/20$\to$14/20. \\
5\x8 $\to$ full 25\x40 patches & two paired $N{=}20$ batches & Across $N{=}40$, kills 60$\to$77 and deaths 33$\to$17, but route mean 5.475$\to$5.225 and descent proxy 40/40$\to$35/40; no completions. \\
100\,ms $\to$ 50\,ms & $N{=}4$ screen & Best route [4,3,3,2], two kills, versus matched production [6,4,4,7], seven kills. \\
Same K=4 sample for 4 passes & $N{=}4$ screen & Held F1 0.588$\to$0.600; selected routes [1,4,6,7], final [3,4,4,4], versus [6,4,4,7]. \\
K=4 $\to$ K=16 distinct windows & $N{=}4$ screen & Held F1 0.588$\to$0.598; best routes [4,5,4,2]. No evidence of live improvement at this sample size. \\
4 $\to$ 8 visual frames & $N{=}4$ screen & Held F1 0.588$\to$0.597; selected routes [2,4,4,7], final [5,4,2,4]. \\
Previous action, 50\% token dropout & $N{=}20$ & Route median 5.5, descent 19/20, kill incidence 17/20, deaths 16; production is 5, 20/20, 19/20, and 15. \\
Failure-similarity sampling & paired $N{=}20$ & Median route 5$\to$6 and deaths 15$\to$10, but mean route 5.45$\to$5.25, descent 20/20$\to$18/20, kill incidence 19/20$\to$16/20. \\
\bottomrule
\end{tabular}
\caption{Retained controlled comparisons. Arrows show the corresponding compact-production measurement, not a claim of statistical significance.}
\label{tab:negative}
\end{table}

\paragraph{Spatial detail changes the trade-off.} The full 25\x40 grid is not a simple regression: it substantially improves combat and survival while slightly reducing opening and route reliability. It also reduces measured training throughput from 2{,}601 to 87.7 examples/s and required an approximately 0.84\,TB packed cache in this implementation. In contrast, batch-one deployed inference rises only from 4.11 to 6.35\,ms, so the large penalty is chiefly in training and packed-data handling. The evidence supports retaining full spatial features for a more efficient multiscale design; it does not support the earlier claim that spatial detail ``does not help.''

\paragraph{Offline metrics are insufficient selectors.} Longer optimization, more distinct windows, longer passive visual context, and denser patches all improve at least one held-out imitation measure without consistently improving route behavior. This is evidence that the measured offline metrics are unreliable selectors in this experiment family. It is not evidence of a general statistical anti-correlation between offline and online performance.

\paragraph{Naive action history is not a free gain.} A previous held-state token makes next-state prediction easy to shortcut. Context dropout reduces the resulting self-lock and yields a competitive route median, but the $N{=}20$ variant does not improve the joint opening, combat, death, and collision profile. This result concerns our simple action-history injection; it does not contradict P2P, whose causal transformer is explicitly trained with past action tokens and correction data.

\paragraph{Recovery-oriented sampling moves behavior.} Weighting windows whose starting features resemble observed failure states reduces deaths in a paired batch but introduces corridor-sprint failures and weakens opening/combat reliability. This is the clearest data-side evidence that recovery and survival are movable, yet it is a trade-off rather than a promoted model.

Across the two production batches, stationary stretches never exceed 3.9 seconds, but 33/40 episodes end in death. Visual audits repeatedly show close-wall and water fixation. Human data contains nearby-looking recovery frames, but demonstrations can disagree about the correct turn. These observations are \emph{consistent with} covariate shift and ambiguous recovery actions; they do not by themselves prove a causal mechanism or prove that DAgger is necessary.

\paragraph{Exploratory visual odometry did not replace engine pose.} Before the retained evaluations, we tested two-frame DINOv3 odometry distilled from a monocular visual-odometry teacher as a game-agnostic progress signal. Successive prototypes mostly learned near-stationary predictions: translation improved only marginally over the stationary prior and rotation remained near it. The teacher labels were noisy; we suspect DINO's semantic invariances and coarse patch grid were a poor match for frame-to-frame correspondence. The resulting signal was not reliable enough for route evaluation, none of the present results uses it, and the exploration is not included as a controlled quantitative result.

\section{Limitations and threats to validity}\label{sec:limits}

No evaluated Cortex episode completes E1M1. The strongest task statement is therefore reliable early-route progress and combat, not level solving. The route score is an order-independent pose heuristic; only engine intermission is success. Door, button-room, and descent regions are not direct interaction events.

The cross-system comparison is not architecture-controlled. Cortex is Quake-specialized; P2P and NitroGen are multi-game policies. They retain different resolutions, histories, action spaces, cadences, and sensitivities. NitroGen additionally depends on our unvalidated gamepad-to-Quake mapping. P2P shares source-data provenance with Cortex, but its training distribution, correction data, and objectives differ. Reference batches contain only five episodes, leaving a 43.4\% upper Wilson bound on a true completion rate after observing 0/5. These results describe released checkpoints under one harness; they do not show that compact specialists generally outperform foundation-scale agents.

The additional maps and save-state starts are exploratory $N{=}4$--5 studies with mixed winners. Many ablations are $N{=}4$ screens. We selected the production checkpoint after iterative experimentation on this environment, so unreported researcher degrees of freedom remain even though the final two batches use fresh seeds. Some raw rollout videos were removed after visual inspection under the project's retention policy; summaries, telemetry, contact sheets, and manifests remain, but a permanent public artifact archive is not yet available.

Finally, total system cost includes the frozen encoder's pretraining and a projected approximately 23 GPU-hours of one-time feature extraction at the measured selective-encoder rate, neither of which appears in the 3.3-minute policy-optimization headline.

The latency comparison likewise describes one RTX 5080 and the eager code paths used by this evaluator. Component boundaries follow each native implementation, preprocessing outside the model is excluded, and NitroGen returns a chunk while the other systems return one decision. The figure therefore compares measured deployed calls, not optimized kernels, equal-horizon control work, energy use, or theoretical compute.

\section{Next steps}\label{sec:future}

\paragraph{Corrective recovery data.} The strongest next experiment is a small DAgger-style dataset~\cite{dagger}: deploy the frozen BC policy, let a human take over in wall, corner, water, and post-combat failure states, and train only on the corrective human segments. P2P reports that less than 1\% correction data mitigates deployment shift~\cite{p2p}. A paired evaluation should test correction data against the existing similarity-weighted approximation.

\paragraph{Efficient full-grid features.} Full 25\x40 DINO features improve combat and survival enough to retain. A multiscale or pooling stem should let every patch contribute without the 30-fold training-throughput loss of flat attention, followed by the same two-batch route/combat evaluation.

\paragraph{Coherent actions and useful memory.} Independent Bernoulli channels can emit implausible chords, and passive history did not help. A compact joint or persistent action decoder should be tested separately from memory. Longer causal memory should be introduced only with an objective that needs it, such as recognizing non-progress or conditioning on a goal.

\paragraph{Universal progress metrics.} The engine pose ladder should eventually be replaced by a game-agnostic measure of progress and stagnation based on visual place recognition, geometric motion, and scene change. The failed DINO odometry prototypes suggest that semantic features alone are insufficient for correspondence; a dedicated geometric representation should be validated against engine state offline, while remaining outside the policy inputs.

\paragraph{Broader evaluation.} Claims about generality require more maps, unseen content, other games, native adapter validation, and larger reference batches. Completion remains the primary endpoint; pose progress is a diagnostic.

\section*{Reproducibility and artifact availability}\label{sec:repro}
\addcontentsline{toc}{section}{Reproducibility and artifact availability}

The evaluated compact policy implementation and action schema are frozen at exact \href{https://github.com/kvark/cortex-actor/tree/b4de4f66420df2c408ec42b5c01c91a088d8b63d}{cortex-actor revision \texttt{b4de4f6}}. The audited results, latency data, and figure generators for this revision are at \href{https://github.com/kvark/cortex-actor/tree/78f78760c3a4b4f145664dd2aa9dd89b553e4d8d}{revision \texttt{78f7876}}; the revised manuscript and arXiv source archive are frozen by the \href{https://github.com/kvark/cortex-actor/tree/paper-v2.2}{\texttt{paper-v2.2} Git tag}. The exact evaluated release checkpoint is at \url{https://huggingface.co/mad-bot/cortex}; its SHA-256 is:
\begin{center}
\small\texttt{29c0e453fdfe7255bc6d8e64a0024fe9b617ed79917f5cd71b41f1173f1aa14b}
\end{center}
Its 86 tensors exactly match the selected step-30{,}000 training checkpoint; only optimizer state, RNG state, and local data paths are removed. The frozen DINOv3 encoder is obtained separately under Meta's license. The public P2P corpus and Quake shareware episode provide the demonstrations and game content used here.

The public release does not contain the full feature-extraction, training, or game-execution pipeline. Evaluation used \href{https://github.com/kvark/vkQuake/tree/def85227e6089231f23c1fbc2ba5e9c454add833}{vkQuake revision \texttt{def85227}}, whose read-only instrumentation exposes pose, health, kills, and intermission state to the evaluator; a separate runtime controlled simulated time. These signals never entered the policy. Cortex targets game-agnostic screen-and-input control, so this Quake adapter is an intermediate evaluation fixture rather than a policy dependency. The evaluator source is archived, but the complete runtime and retained evaluator artifact bundle are not; third-party pixel-to-game reproduction is therefore not yet turnkey.

\section*{Broader impact}

This work studies agents in a 1996 single-player video game. The main transferable risks are those of increasingly capable computer-control agents and of overclaiming results from visually plausible rollouts. Engine-authoritative endpoints, explicit adapter disclosures, and retained visual audits are intended to reduce the latter risk.

The experiment also shows that an individual researcher can run a useful closed-loop game-agent research program on one consumer GPU rather than a server-GPU cluster. This is an accessibility observation, not a claim of compute parity or general state-of-the-art performance: encoder pretraining, public demonstration collection, and one-time feature extraction externalize substantial cost.

\section*{Acknowledgements}

We thank Elefant AI for releasing the Pixels2Play corpus, inference code, and checkpoints; NVIDIA for releasing NitroGen; Axel Gneiting, the author of vkQuake, and the project's contributors; id Software for the Quake engine and shareware episode; and Meta AI for DINOv3.

\section*{AI assistance disclosure}

The model implementation, training and evaluation tooling, experiments, and manuscript were developed with substantial assistance from Anthropic's Claude and OpenAI's Codex under the direction of the author. The author set the goals, reviewed the outputs, and takes responsibility for all claims. Quantitative claims are intended to trace to the artifact ledger accompanying the source.

\appendix
\small

\section{E1M1 route-waypoint heuristic}\label{app:ladder}

For every waypoint independently, the evaluator asks whether the player ever comes within 128 world units. The episode score is the largest index that satisfies this test. It does \emph{not} scan in order and does not require earlier waypoints. The fixed-spawn geometry makes large skips uncommon, but the metric can over-credit an unusual shortcut or teleport. It is a progress diagnostic, never the completion criterion.

\begin{table}[ht]
\centering
\small
\begin{tabular}{llrrr}
\toprule
\# & Landmark & $x$ & $y$ & $z$ \\
\midrule
0 & spawn & 480 & $-352$ & 88 \\
1 & first hall & 480 & 200 & 40 \\
2 & automatic double doors & 232 & 576 & 64 \\
3 & button room & $-64$ & 576 & 24 \\
4 & gate descent & 0 & 576 & $-120$ \\
5 & dark room & 0 & 900 & $-200$ \\
6 & bridge foot & 128 & 1060 & $-200$ \\
7 & bridge north end & 128 & 1400 & $-212$ \\
8 & $y{=}1800$ double doors & 128 & 1800 & $-170$ \\
9 & lift to pool ledge & 544 & 2048 & $-168$ \\
10 & spiral descent & 1000 & 2300 & $-220$ \\
11 & bottom corridor & 1316 & 1128 & $-208$ \\
12 & side door & 1100 & 1024 & $-216$ \\
13 & nailgun bridge & 1312 & 800 & $-204$ \\
14 & exit slipgate & 1312 & 544 & $-204$ \\
\bottomrule
\end{tabular}
\caption{E1M1 reference waypoints, radius 128 world units.}
\label{tab:ladder}
\end{table}

\section{Training details}\label{app:training}

\begin{table}[ht]
\centering
\small
\begin{tabular}{ll}
\toprule
Setting & Value \\
\midrule
Optimizer & AdamW, $\beta_1{=}0.9$, $\beta_2{=}0.95$, weight decay 0.01 \\
Learning rate & $3\times10^{-4}$ cosine-annealed to $3\times10^{-5}$ \\
Gradient clipping & global norm 1.0 \\
Precision & bf16 autocast; cached frozen features in fp8 \\
Maximum batch / steps & 16 / 32{,}320 \\
Loss & active-channel held-state BCE $+ 2.0\x{}$ smooth-L1 mouse loss \\
Sampling & 4 deterministic windows per 128-frame chunk, sampler seed 0 \\
Validation split & 342 held-out recordings \\
Examples / coverage & 517{,}048 / 3.27\% of 15{,}828{,}466 valid train windows \\
Throughput & 2{,}601 examples/s; 198.8\,s optimization time \\
\bottomrule
\end{tabular}
\caption{Production policy training settings.}
\label{tab:training}
\end{table}

The iterable loader assigns chunks to eight workers. Its eight final worker batches are partial (six of size 8 and two of size 4); all other batches contain 16 examples. This accounts for the otherwise non-integral ratio between 517{,}048 examples and 32{,}320 optimizer steps.

\section{Fresh-spawn route arrays}\label{app:episodes}

The complete maximum-waypoint arrays used in Table~\ref{tab:main} are:
\begin{itemize}
\setlength{\itemsep}{0pt}
\setlength{\parsep}{0pt}
\setlength{\parskip}{0pt}
\setlength{\topsep}{2pt}
\setlength{\partopsep}{0pt}
\item Cortex production ($N{=}20$): [7,4,5,4,5,6,4,4,5,6,5,7,6,6,9,6,5,7,4,4].
\item Cortex replication ($N{=}20$): [4,4,6,6,4,7,4,6,7,7,6,4,6,5,6,4,7,6,6,5].
\item P2P-150M ($N{=}5$): [1,1,3,1,2].
\item NitroGen ($N{=}5$): [2,1,2,0,1].
\end{itemize}
All 50 episodes have zero engine-observed completions.

\section{NitroGen gamepad-to-Quake mapping}\label{app:mapping}

This is our adapter, not an official NitroGen Quake configuration. Left-stick Y maps to held \texttt{w}/\texttt{s} and X to \texttt{d}/\texttt{a} beyond a $\pm0.3$ threshold. Right-stick X/Y maps to mouse dx/dy at 20 counts per unit of deflection, SOUTH maps to held fire, and every other button---including jump---is unbound.

\end{document}